\documentclass[letterpaper, 10 pt, conference]{ieeeconf}  % Comment this line out if you need a4paper

\IEEEoverridecommandlockouts                              % This command is only needed if 
\makeatletter
\let\NAT@parse\undefined
\makeatother

\usepackage{cite}
\usepackage{amsmath,amssymb,amsfonts}
\usepackage{algorithmic}
\usepackage{graphicx}
\usepackage{pifont}
\usepackage{textcomp}
\usepackage{hyperref}
\usepackage{multirow}
\usepackage{balance}
\usepackage[table,xcdraw,x11names]{xcolor}
\usepackage{lipsum}
\newcommand{\bx}{{\mathbf x}}
\newcommand{\bX}{{\mathbf X}}
\newcommand{\bA}{{\mathbf A}}

\newcommand{\bU}{{\mathbf U}}

\newcommand{\bQ}{{\mathbf Q}}

\newcommand{\bv}{{\mathbf v}}

\newcommand{\bP}{{\mathbf P}}

\newcommand{\bY}{{\mathbf Y}}
\newcommand{\bg}{{\mathbf g}}
\newcommand{\bB}{{\mathbf B}}
\newcommand{\bu}{{\mathbf u}}

\newcommand{\bz}{{\mathbf z}}

\newcommand{\bJ}{{\mathbf J}}
\newcommand{\bH}{{\mathbf H}}

\newcommand{\bff}{{\mathbf f}}

\newcommand{\mR}{{\mathbb R}}

\newcommand{\cK}{{\cal K}}

\newcommand{\cU}{{\cal U}}

\newcommand{\cX}{{\cal X}}

\newcommand{\bdelta}{{\boldsymbol \delta}}
\newcommand{\bPsi}{{\boldsymbol \Psi}}
\newcommand{\bC}{{\boldsymbol{\mathcal C}}}

\DeclareMathOperator*{\argmin}{arg\,min}

\DeclareMathOperator*{\st}{s.t.}

\title{\LARGE \bf
Digital Twins Meet the Koopman Operator:\\Data-Driven Learning for Robust Autonomy
}

\author{Chinmay V. Samak, Tanmay V. Samak, Ajinkya S. Joglekar, Umesh G. Vaidya and Venkat N. Krovi% <-this % stops a space
\thanks{C.V. Samak, T.V. Samak, A.S. Joglekar and V.N. Krovi are with the Automation, Robotics and Mechatronics Lab (ARMLab), Department of Automotive Engineering, Clemson University International Center for Automotive Research (CU-ICAR), Greenville, SC 29607, USA.
{\tt\small {\{\href{mailto:csamak@clemson.edu}{csamak}, \href{mailto:tsamak@clemson.edu}{tsamak}, \href{mailto:ajoglek@clemson.edu}{ajoglek}, \href{mailto:vkrovi@clemson.edu}{vkrovi}\}@clemson.edu}}}%
\thanks{U.G. Vaidya is with the Dynamics and Control for Autonomy and Intelligence Lab (DyCo AI Lab), Department of Mechanical Engineering, Clemson University, Clemson, SC 29634, USA.
{\tt\small {\href{mailto:uvaidya@clemson.edu}{uvaidya@clemson.edu}}}}%
}

\begin{document}
\maketitle
\thispagestyle{empty}
\pagestyle{empty}

%%%%%%%%%%%%%%%%%%%%%%%%%%%%%%%%%%%%%%%%%%%%%%%%%%%%%%%%%%%%%%%%%%%%%%%%%%%%%%%%

\begin{abstract}

Contrary to on-road autonomous navigation, off-road autonomy is complicated by various factors ranging from sensing challenges to terrain variability. In such a milieu, data-driven approaches have been commonly employed to capture intricate vehicle-environment interactions effectively. However, the success of data-driven methods depends crucially on the quality and quantity of data, which can be compromised by large variability in off-road environments. To address these concerns, we present a novel methodology to recreate the exact vehicle and its target operating conditions digitally for domain-specific data generation. This enables us to effectively model off-road vehicle dynamics from simulation data using the Koopman operator theory, and employ the obtained models for local motion planning and optimal vehicle control. The capabilities of the proposed methodology are demonstrated through an autonomous navigation problem of a 1:5 scale vehicle, where a terrain-informed planner is employed for global mission planning. Results indicate a substantial improvement in off-road navigation performance with the proposed algorithm ($\uparrow5.84\times$) and underscore the efficacy of digital twinning in terms of improving the sample efficiency ($\uparrow3.2\times$) and reducing the sim2real gap ($\downarrow5.2\%$).

\end{abstract}

%%%%%%%%%%%%%%%%%%%%%%%%%%%%%%%%%%%%%%%%%%%%%%%%%%%%%%%%%%%%%%%%%%%%%%%%%%%%%%%%


%%%%%%%%%%%%%%%%%%%%%%%%%%%%%%%%%%%%%%%%%%%%%%%%%%%%%%%%%%%%%%%%%%%%%%%%%%%%%%%%

\section{Introduction}
\label{Section: Introduction}

%%%%%%%%%%%%%%%%%%%%%%%%%%%%%%%%%%%%%%%%%%%%%%%%%%%%%%%%%%%%%%%%%%%%%%%%%%%%%%%%

\begin{figure}[t]
\includegraphics[width=\linewidth]{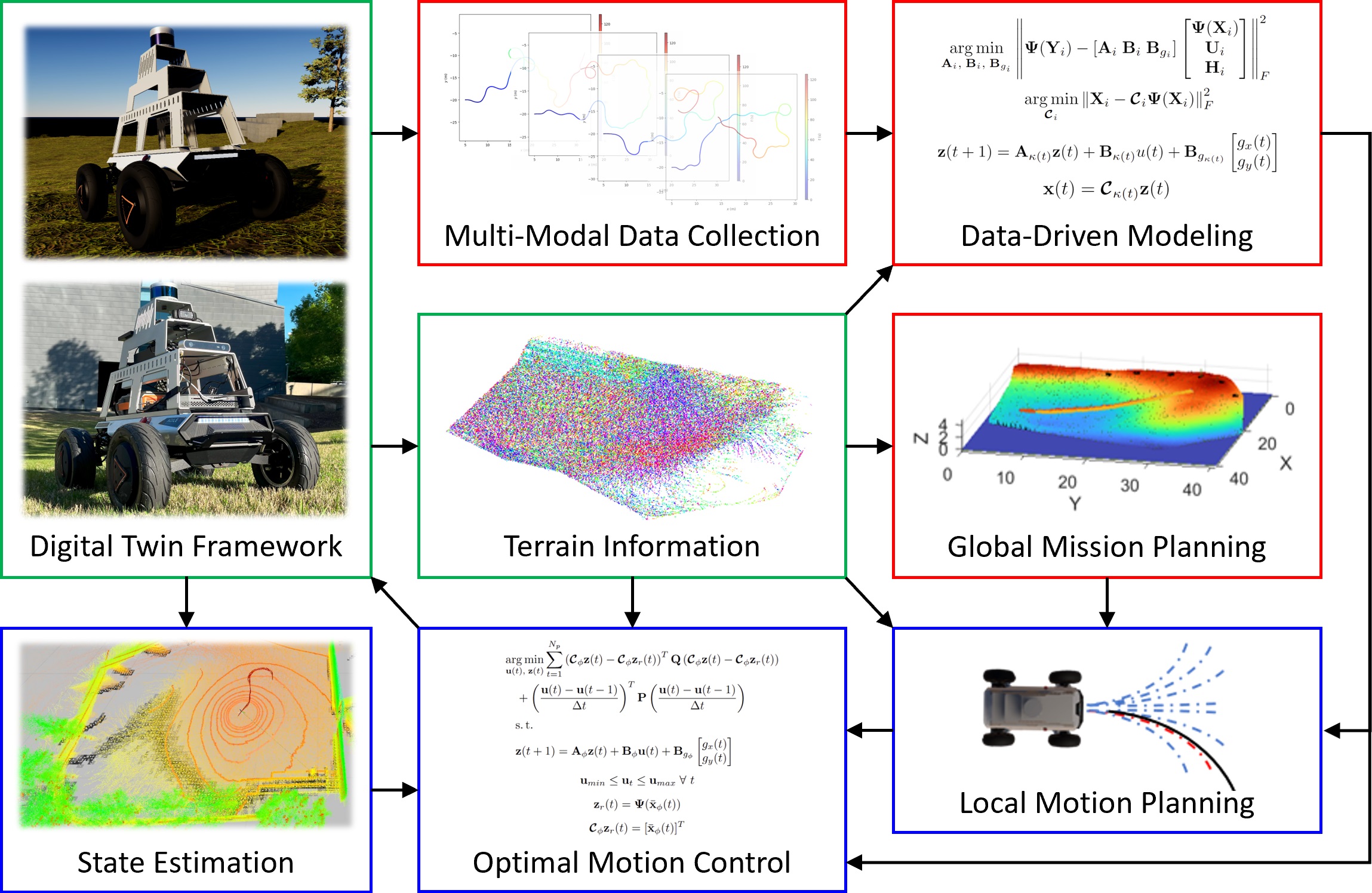}
\caption{The proposed data-driven algorithm for robust off-road autonomous navigation within a digital twin framework. Blocks in {\color{Red1}red} indicate offline processes, {\color{Blue1}blue} ones indicate online processes, and {\color{Green3}green} ones represent items common for online as well as offline processes.}
\label{fig1}
\end{figure}

Autonomous vehicles have revolutionized mobility, with significant advancements in indoor \cite{AMR2011} and on-road autonomy \cite{Levinson2011, Badue2021}. However, off-road environments introduce additional complexity, with unstructured obstacles, uneven terrain, and varying traction conditions, making effective autonomous navigation far more challenging. These scenarios necessitate novel approaches to adequately capture vehicle dynamics \cite{Milliken1995, Rajamani2011} and wheel-terrain interactions \cite{wheel-terrain-2007, wheel-terrain-2012, wheel-terrain-2019}, which are essential for model-based planning and control strategies.

However, the choice of modeling technique plays an important role in addressing the trade-offs concerning accuracy, explainability, simplicity, etc. Traditional first-principles approaches rely on the laws of physics to derive white-box models but generally apply certain simplifying assumptions and approximations to find a reasonable trade-off between modeling accuracy and computational performance. Contrarily, data-driven approaches can potentially capture any unmodeled dynamics, environmental uncertainties, parameter variations, etc. but rely heavily on the quality and quantity of the underlying data, often resulting in grey/black-box non-linear models causing difficulty in interpretation, analysis, and real-time control execution. Although prior works have reported some success working with such models either directly (e.g., non-linear model predictive control \cite{Rick2019, Yu2021, Jonathan2024}) or indirectly (e.g., piece-wise linearization \cite{Benine-Neto2012}, transverse feedback linearization \cite{DSouza2020}, parameter-varying linear models \cite{Katriniok2011}, etc.), our methodology employs the Koopman operator theory \cite{Koopman1931, Mezic2005, Mezic2015, Budivsic2012, han2020deep, huang2022convex, vaidya2023data, vaidya2024} to obtain data-driven globally linear system model(s), perfect for motion planning and optimal control while maintaining computational feasibility.

% Data-driven methods have become increasingly popular for modeling these intricate interactions due to their inherent capability to capture any unmodeled dynamics, environmental uncertainties, and parameter variations.

% However, their effectiveness often depends on the quality and quantity of data - factors that are harder to control in real-world conditions. To mitigate these issues, this work presents a robust framework for off-road autonomous navigation, integrating data-driven modeling and control with reliable data acquisition and validation through digital twin technology (refer Fig. \hyperref[fig1]{\ref*{fig1}}).

This work presents a robust framework for off-road autonomous navigation, integrating data-driven modeling and control with reliable data acquisition and validation through digital twin technology (refer Fig. \hyperref[fig1]{\ref*{fig1}}). The key contributions of this research can be summarized as follows:

\begin{itemize}
    \item A novel method to develop high-fidelity target-specific vehicle and environment digital twins to enable safe and efficient data-collection for terrain-informed vehicle dynamics modeling using the Koopman operator theory.
    \item An end-to-end off-road autonomy pipeline encompassing terrain-aware global mission planning, model-based local motion planning, and optimal control, considering vehicle configuration and terrain properties while using a single sensor (LiDAR) for onboard state estimation.
    \item Extensive benchmarking and validation of digital twins and autonomy algorithm using a 1:5 scale autonomous vehicle, paralleled across simulation and reality.
\end{itemize}

\section{Formulation}
\label{Section: Formulation}

%%%%%%%%%%%%%%%%%%%%%%%%%%%%%%%%%%%%%%%%%%%%%%%%%%%%%%%%%%%%%%%%%%%%%%%%%%%%%%%%

\subsection{Data-Driven Modeling with Koopman Operator}
\label{Sub-Section: Data-Driven Modeling with Koopman Operator}

In this section, we introduce the linear operator theoretic paradigm of data-driven modeling of dynamical systems with the Koopman operator \cite{Koopman1931}. Consider a system characterized by non-linear dynamics given by:
\begin{equation}\label{eqn:3.1}
\bx_{t+1} = \bff(\bx_t,\; \bu_t);\; \bx \in \cX \subseteq \mR^n, \bu \in \cU \subseteq \mR^m
\end{equation}
Here, the function $\bff : (\cX,\; \cU) \rightarrow \cX$ dictates how the system evolves over time. Now consider a non-linear mapping function  $\mathbf{\bg(\bx)}$ that transforms the given state-space into a higher dimensional functional space. In this lifted space, the dynamics are given by:
\begin{equation}\label{eqn:3.2}
[\cK \bg](\bx) = \bg\circ\bff(\bx,\; \bu)
\end{equation}
Here, $\mathcal{K}$ is the linear Koopman operator governing the dynamics evolution in the Hilbert space. The Koopman operator typically is infinite-dimensional as long as the observed state space isn't a finite set. As a result, for practical applications, approaches like generalized Laplace averages (GLA) \cite{mohr2014construction}, dynamic mode decomposition (DMD) \cite{schmid2010dynamic}, and extended dynamic mode decomposition (EDMD) \cite{EDMD2015} are utilized to obtain a finite-dimensional representation of the Koopman operator from data. This work adopts the EDMD technique.

% Introducing a new vector of basis functions transforms our vector space into a higher (theoretically infinite) dimensional function space where the Koopman operator $\mathcal{K}$ acts as a linear composition operator governing the lifted system dynamics using function composition of $\bff$ and $\bg$ as:
% \begin{equation}\label{eqn:3.2}
% [\cK \bg](\bx) = \bg\circ\bff(x,\; u)
% \end{equation}

% For practical purposes, though, we employ the EDMD technique \cite{EDMD2015} to obtain a finite-dimensional approximation of the Koopman operator from data, which approximates the infinite-dimensional function space with a finite-dimensional subspace spanned by the chosen basis functions and can be considered a finite-dimensional Hilbert space.

Consider temporal snapshot data of the system states and control inputs as $\bX = [\bx_0,\; \ldots,\; \bx_{t-1}] \in \mathbb{R}^{n\times t}$, $\bU = [\bu_0,\; \cdots,\; \bu_{t-1}] \in \mathbb{R}^{m\times t}$, with $\bY = [\bx_1,\; \ldots,\; \bx_{t}]\in \mathbb{R}^{n\times t}$ being a single step progression of $\bX$. 
% {\color{red}Ajinkya: This text and  needs to be removed as we cannot state this is in a non-linear format 

% The relationship between elements of $\bX$, $\bY$, and $\bU$ can be formulated as a non-linear system in the vector space as:

% \begin{equation}\label{eqn:3.3}
%     \mathbf{x}_{t+1} = \tilde{\bA} \mathbf{x}_{t} + \tilde{\bB} \bu_{t}
% \end{equation}
% }
Let $\bPsi = [\bPsi_{1}(\bx),\; \ldots,\; \bPsi_{N}(\bx)] \in \mathbb{R}^{N}$, where $N >> n$, be a functional basis vector such that $\bX$ and $\bY$ are lifted along the directions of $\bPsi$ resulting in the following:
\begin{subequations}
\begin{equation}\label{eqn:3.4a}
    \bPsi(\bx_{t+1}) = \bA \bPsi(\bx_{t}) + \bB \bu_{t};\; \bX = \bC \bPsi(\bx_{t})
\end{equation}
\begin{equation}\label{eqn:3.4b}
    \bz_{t+1} = \bA \bz_{t} + \bB \bU;\; \bX = \bC \bz_{t}
\end{equation}
\begin{equation}\label{eqn:3.4c}
    \bz_{t} = \bPsi(\bX);\; \bz_{t+1} = \bPsi(\bY)
\end{equation}
\end{subequations}

Here, $\bA \in \mathbb{R}^{N \times N}$ and $\bB \in \mathbb{R}^{N \times m}$ are the system and control matrices respectively in the function space, while $\bC \in \mathbb{R}^{n \times N}$ is the inverse transform mapping lifted states in high-dimensional function space back to the original states in the low-dimensional vector space.

% {\color{red} Ajinkya: This is kind of repeated in equations 10, so we can omit this part for space saving
% The linear least-squares estimation problems for obtaining $\bA, \bB$ and $\bC$ \cite{KMPC2018} from data are given as:
% \begin{subequations}
% \begin{equation}\label{eqn:3.5a}
%     \argmin_{\bA,\; \bB} \left\| \bz_{t+1} - [\bA\; \bB] \begin{bmatrix} \bz_{t} \\ \bU \end{bmatrix} \right\|_{F}^{2}
% \end{equation}
% \begin{equation}\label{eqn:3.5b}
%     \argmin_{\bC} \left\| \bX - \bC \bz_{t} \right\|_{F}^{2}
% \end{equation}
% \end{subequations}
% }

% The corresponding closed-form analytical solutions to Eq. \ref{eqn:3.5a} and Eq. \ref{eqn:3.5b} are given as:
% \begin{subequations}
% \begin{equation}\label{eqn:3.6a}
%     [\bA\; \bB] = \bPsi(\bY)\begin{bmatrix} \bPsi(\bX) \\ \bU \end{bmatrix}^{\dagger}
% \end{equation}
% \begin{equation}\label{eqn:3.6b}
%     \bC = \bX \begin{bmatrix}\bPsi(\bX)\end{bmatrix}^{\dagger}
% \end{equation}
% \end{subequations}

%%%%%%%%%%%%%%%%%%%%%%%%%%%%%%%%%%%%%%%%%%%%%%%%%%%%%%%%%%%%%%%%%%%%%%%%%%%%%%%%

\subsection{Parameterized Family of Koopman Models}
\label{Sub-Section: Parameterized Family of Koopman Models}

The above section explains data-driven identification of a single global Koopman model based on the snapshots capturing the evolution of a dynamical system subject to control inputs \cite{AjinkyaMECC2023, AjinkyaIROS2023}. However, when applied to systems operating over a wide envelope, a single model may not accurately capture the dynamics and would be susceptible to any biases in the training data. In this context, parameterization of the operating envelope can enable the formulation of multiple models spanning different localized regions of the state-action space, thereby effectively capturing the overall dynamics of the system. These models can then be switched online based on the current operating conditions \cite{MMPK2024}.

In the context of vehicle dynamics application, consider a family of models $\left\{\bA_{\kappa(t)},\; \bB_{\kappa(t)},\; \bC_{\kappa(t)}\right\}$ parameterized by the instantaneous trajectory curvature $\kappa(t) \in [-\kappa_{max},\; \kappa_{max}]$.
For the collected dataset (refer Section \ref{Sub-Section: Multi-Modal Data Collection}), $\kappa_{max}$ = $0.8$ m$^{-1}$, and thus the training data can be split into $q$ = 8 discrete bins as $\{(-0.8,\; -0.6);\; \dots;\; (-0.2,\; 0);\; (0,\; 0.2);\; \dots;\; (0.6,\; 0.8)\}$. During deployment, by considering the instantaneous trajectory curvature, we can identify the associated operating window and opt for the suitable model given by:
\begin{align}\label{eqn:3.7}
    \begin{Bmatrix}\bA_{\kappa(t)} \\ \bB_{\kappa(t)} \\ \bC_{\kappa(t)} \end{Bmatrix} =\left\{
    \begin{array}{ccl}
    \bA_i,\; \bB_i,\; \bC_i & i = 1,\; \ldots,\; q \\
    \bA_1,\; \bB_1,\; \bC_1 & \kappa(t) < -\kappa_{max} \\
    \bA_q,\; \bB_q,\; \bC_q & \kappa(t) > \kappa_{max}
    \end{array}\right.
\end{align}

Thus, the finite-dimensional lifted dynamics for the curvature parameterized family of Koopman models $\forall t \in \mathbb{Z}^+$ can be written as:
\begin{subequations}
\begin{equation}\label{eqn:3.8a}
\bz(t+1)=\bA_{\kappa(t)} \bz(t)+\bB_{\kappa(t)} u(t)
\end{equation}
\begin{equation}\label{eqn:3.8b}
\bx(t)=\bC_{\kappa(t)} \bz(t)
\end{equation}
\end{subequations}

%%%%%%%%%%%%%%%%%%%%%%%%%%%%%%%%%%%%%%%%%%%%%%%%%%%%%%%%%%%%%%%%%%%%%%%%%%%%%%%%

\subsection{Augmentation for Off-Road Conditions}
\label{Sub-Section: Augmentation for Off-Road Conditions}

% In the context of off-road autonomous navigation, it is critical to capture the effects of terrain undulations on vehicle dynamics. In this work, we consider these effects to be external disturbances to the system and counter them through an augmented control matrix.
In this work, we expand the Adaptive MMPK algorithm \cite{Ajinkya_Adaptive_MMPK}, which captures the effect of terrain undulations on vehicle dynamics using IMU-based load transfer quantification. Particularly, we propose to utilize terrain gradient information extracted from the environment digital twin for this purpose, thereby eliminating the need for any additional sensors. To this end, consider a 2D height map $M_H$ of the environment:
\begin{equation}
M_H = \begin{pmatrix}
M_{(0,0)} & M_{(0,1)} & \cdots & M_{(0,v)} \\
M_{(1,0)} & M_{(1,1)} & \cdots & M_{(1,v)} \\
\vdots & \vdots & \ddots & \vdots \\
M_{(u,0)} & M_{(u,1)} & \cdots & M_{(u,v)}
\end{pmatrix}
\end{equation}

For our test environment, $u = 40.78$ m and $v = -40.78$ m. This obtained height map can now be processed to obtain a 2D gradient map $M_G$, where the directional gradients at each grid point are be given by:
\begin{equation}
M_G = \nabla M_H = 
\left\{\begin{matrix}
\frac{\partial M_H}{\partial x} = 0.5*\left[M_{H_{(:,j+1)}} - M_{H_{(:,j-1)}}\right]_{j=1}^{v-1} \\
\frac{\partial M_H}{\partial y} = 0.5*\left[M_{H_{(i+1,:)}} - M_{H_{(i-1,:)}}\right]_{i=1}^{u-1}
\end{matrix}\right.
\end{equation}

\begin{figure*}[t]
\includegraphics[width=\linewidth]{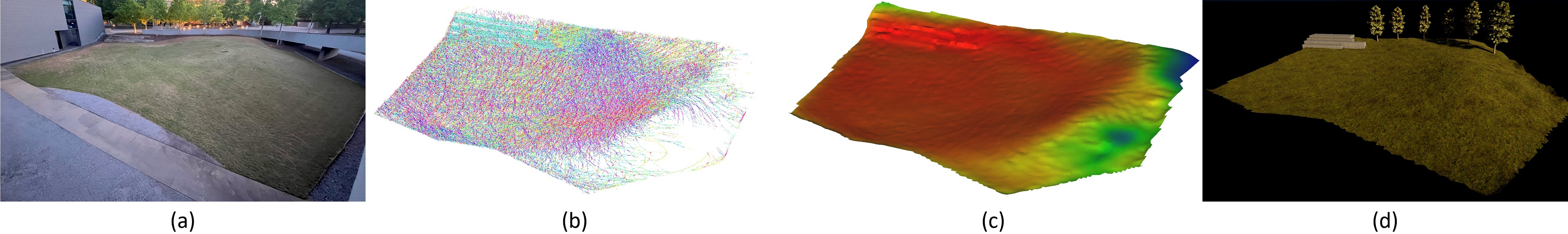}
\caption{Developing digital twin of the off-road greensward area at CU-ICAR within AutoDRIVE Ecosystem: (a) actual environment; (b) LiDAR point cloud data with surface normals; (c) 3D Poisson surface reconstruction; and (d) digital twin of the environment.}
\label{fig3}
\end{figure*}

Now, consider that the training data is partitioned into $q$ discrete curvature bins, denoted as $\bX_i$, $\bY_i$, $\bU_i$ for $i=1,\; \ldots,\; q$. Here, $\bX,\bY$ are the vehicle states describing its pose in polar space one time-step apart such that $\bx(t) = [r,\theta]^T$, and $\bU$ is the control vector consisting of velocity and steering angle inputs $\bu(t) = [v,\delta]^T $. To consider undulation effects, we obtain the temporal snapshots of the terrain gradients by querying the $\nabla M_H$ matrices corresponding to the vehicle pose in $\bX$, mapped to the nearest grid resolution. This returns tuples $(g_x,g_y)$, which contain the localized gradients in $x$ and $y$ directions. Let $\bH \in \mR^{2 \times t} $ be the matrix containing these gradients, such that each element in $\bH$ is given by $\xi(t) = [g_x,g_y]^T$. Similar to the state and control matrices, $\bH$ is parameterized to yield $\bH_i$.

It is assumed that these effects of terrain gradients enter the lifted dynamics affinely. The disturbances from these effects can be mitigated through an augmented control matrix $\bB_{g_{i}}$, changing the lifted dynamics in Eq.\ref{eqn:3.8a} to the following form:
\begin{equation}\label{eqn:3.10}
\bz(t+1)=\bA_{\kappa(t)} \bz(t)+\bB_{\kappa(t)} u(t) + \bB_{g_{\kappa(t)}}    \begin{bmatrix}
        g_x (t) \\
        g_y (t)
    \end{bmatrix}  
\end{equation}

% \begin{equation}\label{eqn:3.9}
%     \bH_i = 
%     \begin{bmatrix}
%         g_{x_i}(0) & g_{y_i}(0) \\
%         g_{x_i}(1) & g_{y_i}(1) \\
%         \vdots & \vdots \\
%         g_{x_i}(n) & g_{y_i}(n)
%     \end{bmatrix};\; i = 1,\; \ldots,\; q
% \end{equation}

% {\color{red} Twins: Why do $x$ and $y$ both have $n$ as last term? Also, aren't both of these supposed to be $m \times n$ 2D matrices, not 1D vectors? We need to figure out how to write this correctly without being confusing.}

The above linear system is achieved through the EDMD technique by following optimization routine:

% synthesizes relevant control commands such that a discrete-time linear system is enabled from the temporal snapshots. This is accomplished by modifying the least-squares problem outlined in Eq. \ref{eqn:3.5a} and Eq. \ref{eqn:3.5b} to the one given below:
\begin{subequations}
\begin{equation}\label{eqn:3.10a}
    \argmin_{ \bA_{i},\; \bB_{i},\; \bB_{g_{i}}} 
    \left\| 
    \bPsi(\bY_i) - 
    \left[ \bA_{i}\;  \bB_{i}\; \bB_{g_{i}} \right]
    \begin{bmatrix}
        \bPsi(\bX_i) \\
        \bU_i \\
        \bH_i
    \end{bmatrix}   
    \right\|^2_{F}
\end{equation}
\begin{equation}\label{eqn:3.10b}
    \argmin_{\bC_{i}} 
    \left\| \bX_i - {\bC_{i}} \bPsi(\bX_i) \right\|^2_{F} 
\end{equation}
\end{subequations}
Here, $\bPsi$ is monomial expansion of the polar pose such that
$\bPsi = [1, r\cos\theta, r\sin\theta, r^2\cos\theta, r^2\sin\theta, r^3\cos\theta, r^3\sin\theta]^T$.

% {\color{red}Ajinkya: This paragraph is confusing and repetitive. I am rewriting it for better flow

% Here, the choice of basis function vector was formulated to be: $\bPsi = [1,\; {\mathrm{d}x}\cos({\mathrm{d}\theta}),\; {\mathrm{d}x}\sin({\mathrm{d}\theta}),\; {\mathrm{d}x}^2\cos({\mathrm{d}\theta}),\;\\ {\mathrm{d}x}^2\sin({\mathrm{d}\theta}),\; {\mathrm{d}x}^3\cos({\mathrm{d}\theta}),\; {\mathrm{d}x}^3\sin({\mathrm{d}\theta})]^T$, with ${\mathrm{d}x}$ being positional displacement and $({\mathrm{d}\theta}$ being orientational displacement of the vehicle in local frame of reference between $\bX_i$ and $\bY_i$ snapshots. Thus, to suit the off-road conditions, we achieve a family of Koopman models for the discretized curvature bins as represented in Eq. \ref{eqn:3.10a} and Eq. \ref{eqn:3.10b}.
% }

%%%%%%%%%%%%%%%%%%%%%%%%%%%%%%%%%%%%%%%%%%%%%%%%%%%%%%%%%%%%%%%%%%%%%%%%%%%%%%%%
\subsection{Global Mission Planning}
\label{Sub-Section: Global Mission Planning}

%%%%%%%%%%%%%%%%%%%%%%%%%%%%%%%%%%%%%%%%%%%%%%%%%%%%%%%%%%%%%%%%%%%%%%%%%%%%%%%%

% In indoor and on-road conditions, global mission planners typically operate under the assumption that the objective space is represented as a two-dimensional Cartesian plane, where specific areas are rendered inaccessible due to the presence of obstacles. However, in the case of off-road vehicles, terrains may introduce elevation changes and excite the roll/pitch dynamics of the vehicle, thereby elevating the complexity of mission planning to a three-dimensional scale. However, optimal planning in such higher-dimensional spaces often translates to increased computational complexity and solution duration. Additionally, the presence of obstacles and untraversable terrain gradients adds to this problem. Thus, we find a middle ground in terms of accuracy and performance by performing terrain-informed global mission planning in a 2.5D objective space leveraging a digital elevation model (DEM) extracted from the digital twin of the environment (refer Fig. \hyperref[fig5]{\ref*{fig5}(a)}).

In this section, we discuss our approach towards terrain-informed global mission planning in a 2.5D objective space leveraging a digital elevation model (DEM) extracted from the digital twin of the environment (refer Section \ref{Sub-Section: Environment Digital Twin}). To this end, we define a synchronized multi-layer map for effectively incorporating the intricate terrain information. This ensures a consistent representation of the Cartesian space across the different layers. The first layer comprises the terrain height map $M_H$. The second layer incorporates the terrain gradients in $x$ and $y$ directions. The third layer holds the gradient costs $C_g = \left\{\begin{matrix}\frac{|g|}{g_{\textrm{max}}} & 0 \leq |g| \leq g_{\textrm{max}}\\e^{\left(\frac{|g|}{g_{\textrm{max}}} - 1\right)} & |g| \geq g_{\textrm{max}} \end{matrix}\right.$ in $x$, $y$ and diagonal directions, where $g$ is the gradient in each direction and $g_{\textrm{max}} = \tan(15^\circ)$. Finally, the fourth layer hosts the obstacle map and gradient-restricted regions considering vehicle actuation limits and traversability constraints.

We employ this DEM to generate safe, optimal, and kinodynamically feasible trajectories for our Ackermann-steered vehicle using Hybrid A* algorithm \cite{Dolgov2008}. To better suit off-road conditions, we define 3 heuristics in addition to the default one to compute the cost associated with transitioning the planning state $s_p = [x_p,y_p]$. The first heuristic uses an elevation-aware cost, $C_{\textrm{EA}} = \|s_2-s_1\| + w*(z_2-z_1)$, which avoids elevations in the terrain with the aim of energy conservation. The second one computes a gradient-aware cost, $C_{\textrm{GA}} = \|s_2-s_1\| + w*|\overrightarrow{s_2-s_1}|*C_g\left<g_x,\; g_y\right>$, which avoids gradients along the heading direction, thereby minimizing pitching for better ride comfort. Finally, the third heuristic uses a rollover-aware cost, $C_{\textrm{RA}} = \|s_2-s_1\| + w*|\overrightarrow{s_2-s_1}|*C_g\left<g_y,\; g_x\right>$, which penalizes terrain gradients perpendicular to the heading direction, thereby minimizing vehicle roll for enhanced safety. Each of these heuristics results in a different mission plan, depending on terrain profile, obstacle positions, and source and goal locations.

%%%%%%%%%%%%%%%%%%%%%%%%%%%%%%%%%%%%%%%%%%%%%%%%%%%%%%%%%%%%%%%%%%%%%%%%%%%%%%%%
\subsection{Local Motion Planning}
\label{Sub-Section: Local Motion Planning}

%%%%%%%%%%%%%%%%%%%%%%%%%%%%%%%%%%%%%%%%%%%%%%%%%%%%%%%%%%%%%%%%%%%%%%%%%%%%%%%%

This section elucidates the model-based local motion planner. Considering Eq. \ref{eqn:3.7}, assume an internal state variable for each of the parameterized models as $\bar{\bx}_{i};\; i = 1,\; \ldots,\; q$ with $\bar{\bx}_{i} =[r\; \theta]^T$ and it's corresponding lifted space variable as $\bar{\bz}_{i};\; i = 1,\; \ldots,\; q$. Now, given the prior control input to the vehicle $\bu(t-1) \in [v(t-1),\; \delta(t-1)]$, by solely examining the velocity $v$ within each curvature bin, we can determine a corresponding steering angle $\delta_{i}$ within the permissible steering range to meet the curvature constraints of the respective bins. Consequently, the effective internal control input variable can be expressed as $\bar{\bu}_{i}\; \in [v(t-1),\; \delta_{i}]$. Thus, for each curvature bin $i \in \{1,\; \ldots,\; q\} $, the internal state $\bar{\bx}_{i}$ can be propagated as follows:
\begin{subequations}
\begin{multline}\label{eqn:5.1a}
    \bar{\bz}_{i}(t \to t+N_{lp}) = \bz(t-1) \\ + \sum_{j=1}^{N_{lp}} \bA_{_i} \bar{\bz}_{i}(t+j)+\bB_{i} \bar{\bu}_{i} + \bB_{g_{i}}\begin{bmatrix} g_x(t-1) \\ g_y(t-1) \end{bmatrix}
\end{multline}
\begin{equation}\label{eqn:5.1b}
     \bar{\bx}_{i}(t \to t+N_{lp}) = \bC_{i}\bar{\bz}_{i}(t \to t+N_{lp})
\end{equation}
\end{subequations}

\begin{figure*}[t]
\includegraphics[width=\linewidth]{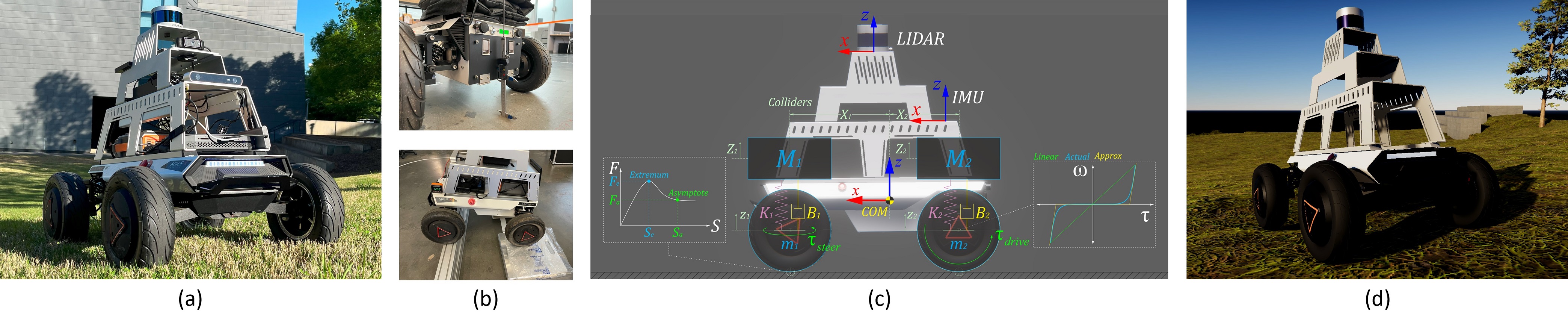}
\caption{Developing digital twin of autonomy-enabled Hunter SE within AutoDRIVE Ecosystem: (a) physical twin of the vehicle; (b) vehicle characterization experiments; (c) simulation models for vehicle dynamics, sensor characteristics, and actuator responses; and (d) digital twin of the vehicle.}
\label{fig2}
\end{figure*}

Thus, we have a predicted trajectory $\bar{\bx}_{i}(t \to t+N_{lp})$ in polar space for each curvature bin $i$, where $N_{lp}$ is the extended prediction horizon used for the local motion planner. The corresponding Cartesian coordinates $[\hat{x}_{i}\; \hat{y}_{i}]^T \in \cX \subseteq \mR^2$ for each of the bins for the internal state $\bar{\bx}_{i}$ can be approximated through the inverse Jacobian transform as $[\hat{x}_{i}\; \hat{y}_{i}]^T  = \bJ^{-1} (\bar{\bx}_{i})$.
These trajectories represent the forward reachable sets (in Cartesian space) rolled out from the current pose of the vehicle, for each of the parameterized Koopman models. The local planner then evaluates which candidate $\phi$ has the least cross-track error with respect to the global mission plan $[x_{p}\; y_{p}]^T \in \cX \subseteq \mR^2$:
\begin{equation}
\label{eqn:5.3}
 \phi = \argmin_{i}
\sqrt{ (\hat{x}_{i} - x_p)^2 +  (\hat{y}_{i} - y_p)^2}
\end{equation}

The local planner outputs the trajectory $\bar{\bx}_{\phi}(t \to t+N_{lp})$ associated with $\phi$ as the reference to a linear model predictive controller with prediction horizon $N_p$ (refer Section \ref{Sub-Section: Optimal Motion Control}).

%%%%%%%%%%%%%%%%%%%%%%%%%%%%%%%%%%%%%%%%%%%%%%%%%%%%%%%%%%%%%%%%%%%%%%%%%%%%%%%%
\subsection{Optimal Motion Control}
\label{Sub-Section: Optimal Motion Control}

%%%%%%%%%%%%%%%%%%%%%%%%%%%%%%%%%%%%%%%%%%%%%%%%%%%%%%%%%%%%%%%%%%%%%%%%%%%%%%%%

In this section, we discuss the linear model predictive control (MPC) law employed for tracking the reference trajectory generated by the local motion planner. Since the linear MPC law can only be applied in the higher dimensional function space, a trivial but necessary step is to lift the reference trajectory as per Eq. \ref{eqn:6.1d}. Assuming we have the optimal curvature bin index $\phi$ from Eq. \ref{eqn:5.3}, we can trace the suitable model from Eq. \ref{eqn:3.7} and solve the optimal control problem given by:
\begin{subequations}\label{eqn:6.1}
\begin{multline}\label{eqn:6.1a}
\argmin_{\bu(t),\; \bz(t)} \sum_{t=1}^{N_p} \left(\bC_{\phi} \bz(t) - \bC_{\phi} \bz_{r}(t)\right)^{T} \bQ \left(\bC_{\phi} \bz(t) - \bC_{\phi} \bz_{r}(t)\right)\\
+ \left(\frac{\bu(t)-\bu(t-1)}{\Delta t}\right)^{T} \bP \left(\frac{\bu(t)-\bu(t-1)}{\Delta t}\right)
\end{multline}
\indent \indent $\st$
\begin{equation}\label{eqn:6.1b}
    \bz(t+1) =\bA_{\phi} \bz(t)+\bB_{\phi} \bu(t) +\bB_{g_{\phi}}    \begin{bmatrix}
        g_x (t) \\
        g_y (t)
    \end{bmatrix}  
\end{equation}
\begin{equation}\label{eqn:6.1c}
     \bu_{min} \leq \bu_t \leq \bu_{max}\; \forall\ t
\end{equation}
\begin{equation}\label{eqn:6.1d}
    \bz_r(t) = \bPsi(\bar{\bx}_{\phi}(t))
\end{equation}
\begin{equation}\label{eqn:6.1e}
    \bC_{\phi} \bz_r(t) = [\bar{\bx}_{\phi}(t)]^T
\end{equation}
\end{subequations}

The optimal solution $\mathbf{u^*} \in \mathbb{R}^{2 \times N_p}$ to the above problem is a vector consisting of longitudinal velocity $\bv$ and steering angle $\bdelta$ commands over a receding horizon, of which, only the first sequence is fed to the autonomous vehicle.

% {\color{red} Ajinkya: redundant 
% Our experiments employ a single 3D LiDAR for map-based localization, which provides live state estimates to be fed back to the local planning and control stages recursively. This eliminates the dependency on any external infrastructure, which particularly suits off-road autonomy (e.g. GNSS-denied areas).
% }

%%%%%%%%%%%%%%%%%%%%%%%%%%%%%%%%%%%%%%%%%%%%%%%%%%%%%%%%%%%%%%%%%%%%%%%%%%%%%%%%
\section{Methodology}
\label{Section: Methodology}

%%%%%%%%%%%%%%%%%%%%%%%%%%%%%%%%%%%%%%%%%%%%%%%%%%%%%%%%%%%%%%%%%%%%%%%%%%%%%%%%

Our work leverages AutoDRIVE Ecosystem \cite{AutoDRIVE-Ecosystem-2023} to develop physically and graphically accurate digital twins of a 1:5 scale autonomous vehicle and its operating environment. This choice was primarily influenced by the simulation fidelity, deployment adaptability and seamless real2sim and sim2real transferability requirements.

\subsection{Vehicle Digital Twin}
\label{Sub-Section: Vehicle Digital Twin}

Our research employs the AgileX Hunter SE (refer Fig. \hyperref[fig2]{\ref*{fig2}(a)}), a 1:5 scale Ackermann-steered, rear-wheel-driven vehicle retrofitted in-house with an Intel NUC 11 Extreme for onboard computation and a Velodyne VLP-16 3D LiDAR for exteroceptive perception. Key geometric, kinematic, static, and dynamic parameters of the physical vehicle were measured/estimated in order to re-create its digital twin using analytical methods (refer Fig. \hyperref[fig2]{\ref*{fig2}(b)}).

This section highlights key aspects of developing vehicle digital twin for the sake of completeness. We direct interested readers to \cite{AutoDRIVE-Autoware-2024} for further details about the modeling and characterization of various autonomous vehicles and their operating environments within AutoDRIVE Simulator \cite{AutoDRIVE-Simulator-2021}.

The vehicle digital twin (refer Fig. \hyperref[fig2]{\ref*{fig2}(c)} and \hyperref[fig2]{\ref*{fig2}(d)}) was modeled using sprung-mass ${^iM}$ and rigid-body representations, which simulates the suspension forces acting on the sprung $M$ and unsprung $m$ masses. The vehicle's electric powertrain was modeled to apply a driving torque ${^i\tau_{drive}}$ to its rear wheels. The torque-speed curve of the driving actuators was characterized to account for any non-linearities and actuation delays. The steering actuator of the vehicle was modeled to apply a steering torque $\tau_{\text{steer}}$ to achieve the desired steering angle $\delta$ at a smooth rate $\dot{\delta} = \kappa_\delta + \kappa_v  \frac{v}{v_\text{max}}$, without exceeding the steering limits $\pm \delta_\text{lim}$. Here, $v$ is the vehicle speed, $\kappa_\delta$ is the steering sensitivity, and $\kappa_v$ is the speed-dependency factor of the steering mechanism. The Ackermann steering constraint was imposed by considering the wheelbase $l$ and track width $w$ of the vehicle:
% $
% \left\{
% \begin{matrix} 
% \delta_{l} = \textup{tan}^{-1}\left(\frac{2l\textup{tan}(\delta)}{2l+w\textup{tan}(\delta)}\right) \\ 
% \delta_r = \textup{tan}^{-1}\left(\frac{2l\textup{tan}(\delta)}{2l-w\textup{tan}(\delta)}\right) 
% \end{matrix}
% \right.
% $
$\delta_{l/r} = \textup{tan}^{-1}\left(\frac{2l\textup{tan}(\delta)}{2l \pm w\textup{tan}(\delta)}\right)$.

Tire forces were simulated by querying the friction characteristics of $i$-th tire based on the longitudinal $^iS_x$ and lateral $^iS_y$ slip values. Here, the friction characteristics are approximated using a two-piece spline defined as $F(S) = \left\{\begin{matrix} f_0(S); \;\; S_0 \leq S < S_e \\ f_1(S); \;\; S_e \leq S < S_a \\ \end{matrix}\right.$, with $f_k(S)$ representing a cubic polynomial function. The first segment of the spline extends from the origin $(S_0,F_0)$ to an extremum point $(S_e,F_e)$, while the second segment extends from the extremum point $(S_e, F_e)$ to an asymptote point $(S_a, F_a)$.

%%%%%%%%%%%%%%%%%%%%%%%%%%%%%%%%%%%%%%%%%%%%%%%%%%%%%%%%%%%%%%%%%%%%%%%%%%%%%%%%

\subsection{Environment Digital Twin}
\label{Sub-Section: Environment Digital Twin}

\begin{figure*}[t]
\includegraphics[width=\linewidth]{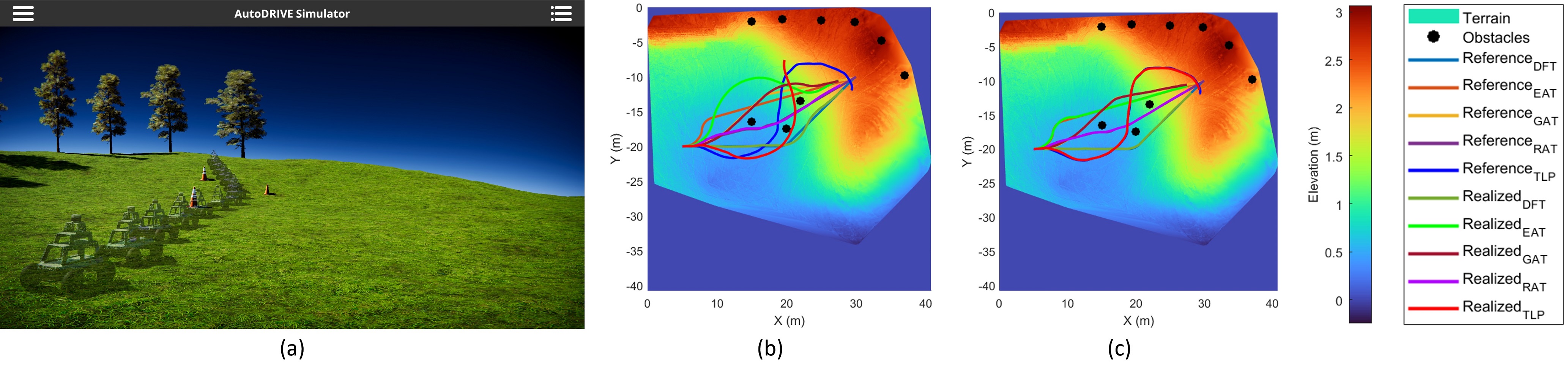}
\caption{Benchmarking and validation of the proposed autonomy algorithm in simulation: (a) freeze-frame sequence sampled at 0.5 Hz depicting execution of the rollover-aware mission plan in AutoDRIVE Simulator; (b) tracking performance of the non-augmented algorithm for optimally planned and aggressively teleoperated trajectories; and (c) tracking performance of the augmented algorithm for optimally planned and aggressively teleoperated trajectories.}
\label{fig6}
\end{figure*}

While previous environment modeling efforts with AutoDRIVE Ecosystem have predominantly focused on multi-scale synthetic scenarios, this work presents the first-ever sub-millimeter-level accurate ($\mu_{dist}$ = 2.28e-4 m) digital twin representation of an actual off-road greensward area at CU-ICAR (refer Fig. \hyperref[fig3]{\ref*{fig3}(a)}). To this end, we employed the physical Hunter SE to collect high-resolution point cloud data of this area using its 3D LiDAR (from a single experimental run). This data was then post-processed to remove any unwanted artifacts by segmenting it. This was followed by calculating the surface normals and aligning them (refer Fig. \hyperref[fig3]{\ref*{fig3}(b)}). We then performed 3D Poisson surface reconstruction \cite{Kazhdan2006} over the aligned point cloud data with an octree depth of 7 (refer Fig. \hyperref[fig3]{\ref*{fig3}(c)}). Finally, the terrain was processed with physical (e.g., collider, friction map, etc.) and graphical (e.g., materials, textures, etc.) properties, along with sim-ready assets such as trees and steps (refer Fig. \hyperref[fig3]{\ref*{fig3}(d)}).

In addition to embedding the resultant digital twin environment within AutoDRIVE Simulator for supporting high-quality data collection and simulation-based validation of our off-road navigation algorithm, we also extracted meaningful terrain information for run-time algorithmic use (refer Section \ref{Sub-Section: Augmentation for Off-Road Conditions}). Particularly, a finely sampled (0.02 m resolution in our case) 2.5D digital elevation model of the environment was synthesized from its 3D digital twin, which was further down-sampled (0.5 m resolution in our case, considering wheelbase of our test vehicle -- refer Section \ref{Sub-Section: Vehicle Digital Twin}) to generate a 2D height map of the environment.

%%%%%%%%%%%%%%%%%%%%%%%%%%%%%%%%%%%%%%%%%%%%%%%%%%%%%%%%%%%%%%%%%%%%%%%%%%%%%%%%

\subsection{Multi-Modal Data Collection}
\label{Sub-Section: Multi-Modal Data Collection}

% \begin{figure*}[t]
% \includegraphics[width=\linewidth]{Figures/Figure_4.jpg}
% \caption{Multi-modal data collection within a synthetic off-road environment: (a) virtual environment used for data collection; (b) keyboard teleoperation; (c) mouse teleoperation; (d) gamepad teleoperation; and (e) driving rig teleoperation.}
% \label{fig4}
% \end{figure*}
This section describes the process of multi-modal data collection, which served as the basis for data-driven modeling (refer Section \ref{Sub-Section: Data-Driven Modeling with Koopman Operator}). Since there were no established benchmarks for systematically exciting off-road vehicles (due to the inherent environmental variability), we teleoperated the vehicle to recorded time-synchronized data comprising its states $\{n_l,n_r,x,y,z,R,P,Y,v,\omega_x,\omega_y,\omega_z,a_x,a_y,a_z\}$ and corresponding control inputs $\{\tau,\delta\}$ at 30 Hz, while ensuring $>75\%$ feature coverage. Here, $n_l$ and $n_r$ are the left and right encoder ticks, $\{x,y,z\}$ are the positional coordinates, $\{R,P,Y\}$ describe the orientation (Euler angles), $v$ is the speed, $\{\omega_x,\omega_y,\omega_z\}$ are the angular velocities, $\{a_x,a_y,a_z\}$ are the linear accelerations, and $\{\tau,\delta\}$ are the throttle and steering commands of the vehicle.

% The dataset was recorded in 4 batches, each with a different modality, viz. keyboard (binary-resolution), mouse (high-resolution), gamepad (low-resolution) and driving rig (extremely high-resolution). Two individuals collected the dataset (person-1: keyboard and mouse teleoperation, person-2: gamepad and driving rig teleoperation) to avoid any biases in the driving behavior. Each data batch ran 5 distinct throttle gradations $\tau \in \{0.1, 0.2, 0.3, 0.4, 0.5\}$ norm\%, with each run lasting approximately 120 seconds. This resulted in a total of 20 runs amounting over 40 minutes of driving data.
% The dataset was intentionally kept small to avoid overfitting to the operating conditions.

AutoDRIVE Simulator allowed us to systematically collect vehicle data across different environments while enabling data integrity (quality and quantity) trade studies (refer Section \ref{Sub-Section: Effect of the Digital Twin Framework}). Particularly, we recorded 2 datasets in two different environments. First, we recorded $\sim$40 minutes of driving data from the greensward environment digital twin. Next, we collected $\sim$168 minutes of driving data from a synthetic off-road environment. Each dataset comprised 4 batches, where each batch was recorded with a different modality, viz. keyboard (binary-resolution), mouse (high-resolution), gamepad (low-resolution) and driving rig (extremely high-resolution). Two individuals collected the datasets (person-1: keyboard and mouse teleoperation, person-2: gamepad and driving rig teleoperation) to avoid any biases in the driving behavior. Each data batch ran 5 distinct throttle gradations $\tau \in \{0.1, 0.2, 0.3, 0.4, 0.5\}$ norm\%.

%%%%%%%%%%%%%%%%%%%%%%%%%%%%%%%%%%%%%%%%%%%%%%%%%%%%%%%%%%%%%%%%%%%%%%%%%%%%%%%%
\section{Results and Discussion}
\label{Section: Results and Discussion}

%%%%%%%%%%%%%%%%%%%%%%%%%%%%%%%%%%%%%%%%%%%%%%%%%%%%%%%%%%%%%%%%%%%%%%%%%%%%%%%%

\begin{figure*}[t]
\includegraphics[width=\linewidth]{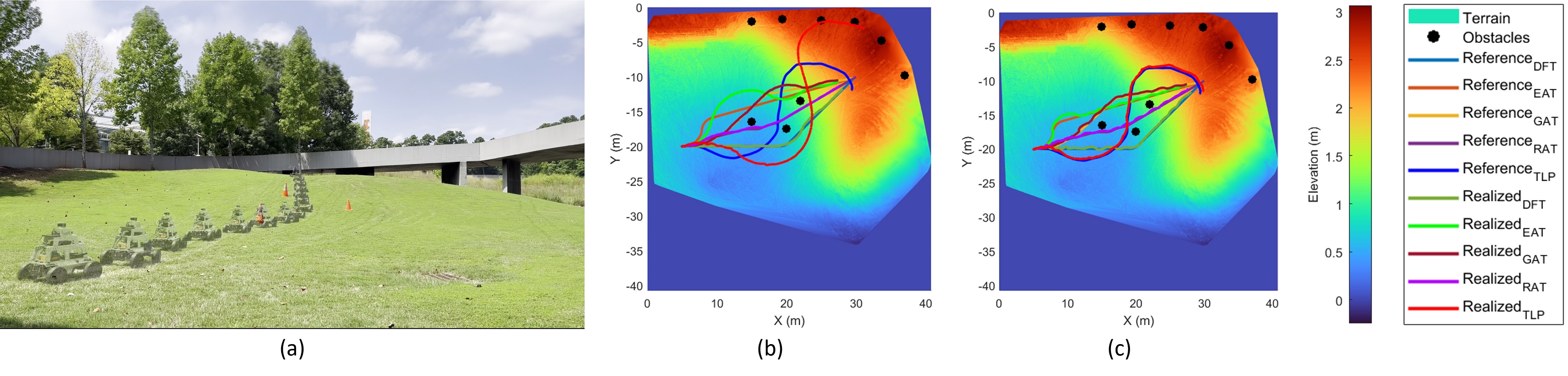}
\caption{Benchmarking and validation of the proposed autonomy algorithm in the real world: (a) freeze-frame sequence sampled at 0.5 Hz depicting execution of the rollover-aware mission plan at CU-ICAR; (b) tracking performance of the non-augmented algorithm for optimally planned and aggressively teleoperated trajectories; and (c) tracking performance of the augmented algorithm for optimally planned and aggressively teleoperated trajectories.}
\label{fig7}
\end{figure*}

% \subsection{Design of Experiments}

% The design of experiments was carefully considered to serve a dual purpose: (a) analyzing the effect of digital twinning on the sample efficiency and sim2real transferability of the proposed autonomy algorithm, and (b) validating the proposed autonomy algorithm and benchmarking it against its non-augmented form \cite{MMPK2024}. Findings are reported considering the root mean squared error ($\varepsilon$) in tracking as the key performance indicator.

% analyzing the performance and correlation of the experiments conducted in simulation and the real world to assess the effectiveness of the presented digital twin framework in bridging the sim2real gap. Table \ref{tab1} presents the key performance indicators across all the experiments.

% Consequently, we assess the off-road navigation performance of Hunter SE using the proposed algorithm across optimally planned and aggressively teleoperated trajectories in simulation and real-world experiments.
% Here, the teleoperated reference trajectories can be considered analogous to optimal trajectories generated as a result of non-traversability imposed by obstacles (e.g. trees and boulders), gradients (e.g. crests and troughs), natural elements (e.g. waterbodies), etc.
% The proposed approach is also benchmarked against its non-augmented form, which was originally proposed for on-road autonomous navigation. Table \ref{tab1} presents the key performance indices across all the experiments.

\subsection{Effect of the Digital Twin Framework}
\label{Sub-Section: Effect of the Digital Twin Framework}

First, we analyzed the effect of adopting the exact digital twin of the operating environment, as opposed to using a general synthetic off-road environment for data collection. Here, (a) the synthetic environment was much larger and comprised richer features than the digital twin environment, (b) we used the same vehicle digital twin in both cases, and (c) the global missions were planned without considering any obstacles. Table \ref{tab1} presents the root mean squared error (RMSE), $\varepsilon$, in tracking as the key performance indicator (KPI) to assess the sample efficiency and sim2real gap of the proposed algorithm trained on different datasets.

While both models were able to track the default (DFT), elevation-aware (EAT), gradient-aware (GAT), and rollover-aware (RAT) trajectories generated by the global mission planner quite well, it was observed that the performance of the model trained on the digital twin data was, on average, 12.1\% better across simulated and real-world experiments. This validated the superiority of the digital twin data in terms of increasing the sample efficiency by over 3.2 folds.

It was also observed that the error profiles between simulated and real-world experiments were closer in the case where digital twin data was used. This was marked by a 5.2\% reduction in the difference across the two tracking error values (sim vs. real), which signifies the sim2real gap.

\subsection{Effect of the Augmented Autonomy Algorithm}
\label{Sub-Section: Effect of the Augmented Autonomy Algorithm}

\begin{table}[t]
\centering
\caption{Benchmarking and validation of the digital twin framework.}
\label{tab1}
\resizebox{\columnwidth}{!}{%
\begin{tabular}{l|lll|lll|ll}
\hline
\multirow{2}{*}{\textbf{$\varepsilon$ (m) $\searrow$}} & \multicolumn{3}{c|}{\textbf{Synthetic Data}} & \multicolumn{3}{c|}{\textbf{Digital Twin Data}} & \multicolumn{2}{c}{\textbf{Improvement (\%)}} \\ \cline{2-9} 
& \multicolumn{1}{c|}{Sim}  & \multicolumn{1}{c|}{Real} & \multicolumn{1}{c|}{Gap}  & \multicolumn{1}{c|}{Sim}  & \multicolumn{1}{c|}{Real} & \multicolumn{1}{c|}{Gap}  & \multicolumn{1}{c|}{Sim}  & \multicolumn{1}{c}{Real}  \\ \hline
DFT* & \multicolumn{1}{c|}{0.0615}  & \multicolumn{1}{c|}{0.2155}   & 0.1539  & \multicolumn{1}{c|}{0.0599} & \multicolumn{1}{c|}{0.2057} & 0.1458 & \multicolumn{1}{c|}{2.70e+1}   & 4.74e+1  \\
EAT* & \multicolumn{1}{c|}{0.0640}  & \multicolumn{1}{c|}{0.2224}   & 0.1584  & \multicolumn{1}{c|}{0.0551} & \multicolumn{1}{c|}{0.2047} & 0.1496 & \multicolumn{1}{c|}{1.61e+2}   & 8.62e+1  \\
GAT* & \multicolumn{1}{c|}{0.0650}  & \multicolumn{1}{c|}{0.2126}   & 0.1476  & \multicolumn{1}{c|}{0.0510} & \multicolumn{1}{c|}{0.1890} & 0.1380 & \multicolumn{1}{c|}{2.73e+2}   & 1.25e+2  \\
RAT* & \multicolumn{1}{c|}{0.0572}  & \multicolumn{1}{c|}{0.2107}   & 0.1535  & \multicolumn{1}{c|}{0.0483} & \multicolumn{1}{c|}{0.1983} & 0.1500 & \multicolumn{1}{c|}{1.84e+2}   & 6.26e+1  \\ \hline
\multicolumn{9}{l}{*Global mission plans with the same source and destination, but without any obstacles.} \\
\end{tabular}%
}
\end{table}

After establishing the importance of adopting digital twins, we analyzed the effect of using an augmented control matrix for disturbance rejection, as opposed to not using one \cite{MMPK2024}. Here, (a) we used the same vehicle and environment digital twins in both cases, (b) we used the same training data ($\sim$40 minutes) for both the models, and (c) the global missions were planned for the same source and destination, but with 3 obstacles along the way. Table \ref{tab2} presents the tracking RMSE ($\varepsilon$) of the augmented and non-augmented algorithms.

While both the approaches very well tracked the global mission plans (DFT, EAT, GAT, and RAT), the augmented approach slightly outperformed the non-augmented one in simulation (refer Fig. \ref{fig6}) as well as real-world (refer Fig. \ref{fig7}) experiments. It is to be noted that the non-augmented approach was able to track these trajectories since they did not disturb the vehicle dynamics significantly. This validated the effectiveness of the terrain-aware global mission planner in terms of generating efficient trajectories.

More generically though, it was observed that the non-augmented algorithm was sluggish compared to the slightly aggressive augmented approach, and therefore accumulated a higher error. This hypothesis was stress-tested by selecting an aggressively teleoperated trajectory (TLP), which was never encountered by the model during training. The said TLP trajectory captured higher curvatures and traversed over larger terrain gradients. Owing to its sluggish nature, the non-augmented approach failed to track the TLP trajectory in simulation as well as real-world experiments to an extent where human intervention became necessary. However, it is worth mentioning that the failure profile was almost identical during simulated and physical deployments, which highlighted the reliability of the digital twin framework in predicting or reproducing experimental outcomes. On the other hand, the augmented approach was able to successfully track the TLP trajectory in simulation as well as real-world experiments. This verified the underlying formulation of the augmented approach, which explicitly captured the terrain gradients responsible for disturbing the vehicle dynamics. Additionally, the similarity of tracking performance between simulated and physical experiments can be observed even in this case, which indicates a strong correlation between results obtained from simulation and real-world experiments.

In summary, the augmented approach performed 5.84$\times$ better than the non-augmented one across all trajectories.

\begin{table}[t]
\centering
\caption{Benchmarking and validation of the autonomy algorithm.}
\label{tab2}
\resizebox{\columnwidth}{!}{%
\begin{tabular}{l|ll|ll|ll}
\hline
\multirow{2}{*}{\textbf{$\varepsilon$ (m) $\searrow$}} & \multicolumn{2}{c|}{\textbf{Non-Augmented}} & \multicolumn{2}{c|}{\textbf{Augmented}} & \multicolumn{2}{c}{\textbf{Improvement (\%)}} \\ \cline{2-7} 
& \multicolumn{1}{c|}{Sim}  & \multicolumn{1}{c|}{Real}  & \multicolumn{1}{c|}{Sim}  & \multicolumn{1}{c|}{Real}  & \multicolumn{1}{c|}{Sim}  & \multicolumn{1}{c}{Real}  \\ \hline
DFT & \multicolumn{1}{c|}{0.1972}  & 0.1435                        & \multicolumn{1}{c|}{0.0555}  & 0.1183 & \multicolumn{1}{c|}{2.55e+2} & 2.13e+1  \\
EAT & \multicolumn{1}{c|}{1.8438}  & 1.0258                        & \multicolumn{1}{c|}{0.0780}  & 0.2350 & \multicolumn{1}{c|}{2.26e+3} & 3.36e+2  \\
GAT & \multicolumn{1}{c|}{0.3029}  & 0.2488                        & \multicolumn{1}{c|}{0.0539}  & 0.1355 & \multicolumn{1}{c|}{4.61e+2} & 8.35e+1  \\
RAT & \multicolumn{1}{c|}{0.1132}  & 0.1505                        & \multicolumn{1}{c|}{0.0530}  & 0.1061 & \multicolumn{1}{c|}{1.14e+2} & 4.18e+1  \\
TLP & \multicolumn{1}{c|}{1.2879}  & 4.5391                        & \multicolumn{1}{c|}{0.0923}  & 0.2275 & \multicolumn{1}{c|}{1.29e+3} & 6.15e+3  \\ \hline
\end{tabular}%
}
\end{table}

%%%%%%%%%%%%%%%%%%%%%%%%%%%%%%%%%%%%%%%%%%%%%%%%%%%%%%%%%%%%%%%%%%%%%%%%%%%%%%%%
\section{Conclusion}
\label{Section: Conclusion}

%%%%%%%%%%%%%%%%%%%%%%%%%%%%%%%%%%%%%%%%%%%%%%%%%%%%%%%%%%%%%%%%%%%%%%%%%%%%%%%%

This work presented our attempt to address the critical challenge of off-road autonomous navigation by leveraging the Koopman operator theory and digital twin technology. We introduced a novel methodology to develop target-specific vehicle and environment digital twins, which enabled off-road vehicle dynamics modeling from simulation-only data using the Koopman operator theory. We applied the derived models to formulate an end-to-end off-road autonomy pipeline, starting with terrain-aware global mission planning, followed by model-based local motion planning and optimal control, while using a single sensor (LiDAR) for onboard state estimation. We benchmarked and validated the performance of a 1:5 scale autonomous vehicle using the proposed methodology across multiple experiments in simulation and reality. Results indicated a substantial improvement ($\uparrow5.84\times$) in off-road navigation performance with the proposed algorithm and underscored the efficacy of the digital twin framework in terms of improving the sample efficiency ($\uparrow3.2\times$) and reducing the sim2real gap ($\downarrow5.2\%$).

Future research could study the effect of multi-scale, multi-fidelity simulations on data generation, and explore alternative formulations to Koopman modeling for improving the reliability and versatility of off-road autonomy.

%%%%%%%%%%%%%%%%%%%%%%%%%%%%%%%%%%%%%%%%%%%%%%%%%%%%%%%%%%%%%%%%%%%%%%%%%%%%%%%%

%%%%%%%%%%%%%%%%%%%%%%%%%%%%%%%%%%%%%%%%%%%%%%%%%%%%%%%%%%%%%%%%%%%%%%%%%%%%%%%%

\balance
\bibliographystyle{IEEEtran}
\bibliography{References}

\begin{thebibliography}{10}
\providecommand{\url}[1]{#1}
\csname url@rmstyle\endcsname
\providecommand{\newblock}{\relax}
\providecommand{\bibinfo}[2]{#2}
\providecommand\BIBentrySTDinterwordspacing{\spaceskip=0pt\relax}
\providecommand\BIBentryALTinterwordstretchfactor{4}
\providecommand\BIBentryALTinterwordspacing{\spaceskip=\fontdimen2\font plus
\BIBentryALTinterwordstretchfactor\fontdimen3\font minus \fontdimen4\font\relax}
\providecommand\BIBforeignlanguage[2]{{%
\expandafter\ifx\csname l@#1\endcsname\relax
\typeout{** WARNING: IEEEtran.bst: No hyphenation pattern has been}%
\typeout{** loaded for the language `#1'. Using the pattern for}%
\typeout{** the default language instead.}%
\else
\language=\csname l@#1\endcsname
\fi
#2}}

\bibitem{AMR2011}
R.~Siegwart, I.~R. Nourbakhsh, and D.~Scaramuzza, \emph{{Introduction to Autonomous Mobile Robots}}, 2nd~ed.\hskip 1em plus 0.5em minus 0.4em\relax The MIT Press, 2011.

\bibitem{Levinson2011}
\BIBentryALTinterwordspacing
J.~Levinson, J.~Askeland, J.~Becker, J.~Dolson, D.~Held, S.~Kammel, J.~Z. Kolter, D.~Langer, O.~Pink, V.~Pratt, M.~Sokolsky, G.~Stanek, D.~Stavens, A.~Teichman, M.~Werling, and S.~Thrun, ``{Towards Fully Autonomous Driving: Systems and Algorithms},'' in \emph{2011 IEEE Intelligent Vehicles Symposium (IV)}, 2011, pp. 163--168. [Online]. Available: \url{https://doi.org/10.1109/IVS.2011.5940562}
\BIBentrySTDinterwordspacing

\bibitem{Badue2021}
\BIBentryALTinterwordspacing
C.~Badue, R.~Guidolini, R.~V. Carneiro, P.~Azevedo, V.~B. Cardoso, A.~Forechi, L.~Jesus, R.~Berriel, T.~M. Paixão, F.~Mutz, L.~{de Paula Veronese}, T.~Oliveira-Santos, and A.~F. {De Souza}, ``{Self-Driving Cars: A Survey},'' \emph{Expert Systems with Applications}, vol. 165, p. 113816, 2021. [Online]. Available: \url{https://doi.org/10.1016/j.eswa.2020.113816}
\BIBentrySTDinterwordspacing

\bibitem{Milliken1995}
\BIBentryALTinterwordspacing
W.~Milliken and D.~Milliken, \emph{{Race Car Vehicle Dynamics}}, ser. Premiere Series.\hskip 1em plus 0.5em minus 0.4em\relax SAE International, 1995. [Online]. Available: \url{https://books.google.com/books?id=EOHPjgEACAAJ}
\BIBentrySTDinterwordspacing

\bibitem{Rajamani2011}
\BIBentryALTinterwordspacing
R.~Rajamani, \emph{{Vehicle Dynamics and Control}}, ser. Mechanical Engineering Series.\hskip 1em plus 0.5em minus 0.4em\relax Springer US, 2011. [Online]. Available: \url{https://books.google.com/books?id=q6SJcgAACAAJ}
\BIBentrySTDinterwordspacing

\bibitem{wheel-terrain-2007}
T.~H. Tran, N.~M. Kwok, S.~Scheding, and Q.~P. Ha, ``{Dynamic Modelling of Wheel-Terrain Interaction of a UGV},'' in \emph{2007 IEEE International Conference on Automation Science and Engineering}, 2007, pp. 369--374.

\bibitem{wheel-terrain-2012}
Z.~Jia, W.~Smith, and H.~Peng, ``{Terramechanics-Based Wheel–Terrain Interaction Model and its Applications to Off-Road Wheeled Mobile Robots},'' \emph{Robotica}, vol.~30, no.~3, p. 491–503, 2012.

\bibitem{wheel-terrain-2019}
R.~Serban, D.~Negrut, A.~Recuero, and P.~Jayakumar, ``{An Integrated Framework for High-Performance, High-Fidelity Simulation of Ground Vehicle-Tyre-Terrain Interaction},'' \emph{International Journal of Vehicle Performance}, vol.~5, no.~3, pp. 233--259, 2019.

\bibitem{Rick2019}
\BIBentryALTinterwordspacing
M.~Rick, J.~Clemens, L.~Sommer, A.~Folkers, K.~Schill, and C.~Büskens, ``{Autonomous Driving Based on Nonlinear Model Predictive Control and Multi-Sensor Fusion},'' \emph{IFAC-PapersOnLine}, vol.~52, no.~8, pp. 182--187, 2019, 10th IFAC Symposium on Intelligent Autonomous Vehicles IAV 2019. [Online]. Available: \url{https://www.sciencedirect.com/science/article/pii/S2405896319303994}
\BIBentrySTDinterwordspacing

\bibitem{Yu2021}
\BIBentryALTinterwordspacing
S.~Yu, C.~Shen, and T.~Ersal, ``{Nonlinear Model Predictive Planning and Control for High-Speed Autonomous Vehicles on 3D Terrains},'' \emph{IFAC-PapersOnLine}, vol.~54, no.~20, pp. 412--417, 2021, modeling, Estimation and Control Conference MECC 2021. [Online]. Available: \url{https://www.sciencedirect.com/science/article/pii/S2405896321022527}
\BIBentrySTDinterwordspacing

\bibitem{Jonathan2024}
\BIBentryALTinterwordspacing
J.~D. Jonathan Y. M.~Goh, Michael~Thompson and A.~Balachandran, ``{Beyond the Stable Handling Limits: Nonlinear Model Predictive Control for Highly Transient Autonomous Drifting},'' \emph{Vehicle System Dynamics}, vol.~0, no.~0, pp. 1--24, 2024. [Online]. Available: \url{https://doi.org/10.1080/00423114.2023.2297799}
\BIBentrySTDinterwordspacing

\bibitem{Benine-Neto2012}
\BIBentryALTinterwordspacing
A.~Benine-Neto and C.~Grand, ``{Piecewise Affine Control for Fast Unmanned Ground Vehicles},'' in \emph{2012 IEEE/RSJ International Conference on Intelligent Robots and Systems}, 2012, pp. 3673--3678. [Online]. Available: \url{https://doi.org/10.1109/IROS.2012.6385675}
\BIBentrySTDinterwordspacing

\bibitem{DSouza2020}
\BIBentryALTinterwordspacing
R.~S. D’Souza and C.~Nielsen, ``{Piecewise-Linear Path Following for a Unicycle using Transverse Feedback Linearization},'' in \emph{2020 American Control Conference (ACC)}, 2020, pp. 5250--5255. [Online]. Available: \url{https://doi.org/10.23919/ACC45564.2020.9147473}
\BIBentrySTDinterwordspacing

\bibitem{Katriniok2011}
\BIBentryALTinterwordspacing
A.~Katriniok and D.~Abel, ``{LTV-MPC Approach for Lateral Vehicle Guidance by Front Steering at the Limits of Vehicle Dynamics},'' in \emph{2011 50th IEEE Conference on Decision and Control and European Control Conference}, 2011, pp. 6828--6833. [Online]. Available: \url{https://doi.org/10.1109/CDC.2011.6161257}
\BIBentrySTDinterwordspacing

\bibitem{Koopman1931}
\BIBentryALTinterwordspacing
B.~O. Koopman, ``{Hamiltonian Systems and Transformation in Hilbert Space},'' \emph{Proceedings of the National Academy of Sciences}, vol.~17, no.~5, pp. 315--318, 1931. [Online]. Available: \url{https://doi.org/10.1073/pnas.17.5.315}
\BIBentrySTDinterwordspacing

\bibitem{Mezic2005}
\BIBentryALTinterwordspacing
I.~Mezi{\'c}, ``{Spectral Properties of Dynamical Systems, Model Reduction and Decompositions},'' \emph{Nonlinear Dynamics}, vol.~41, pp. 309--325, 2005. [Online]. Available: \url{https://doi.org/10.1007/s11071-005-2824-x}
\BIBentrySTDinterwordspacing

\bibitem{Mezic2015}
\BIBentryALTinterwordspacing
------, ``{On Applications of the Spectral Theory of the Koopman Operator in Dynamical Systems and Control Theory},'' in \emph{2015 54th IEEE Conference on Decision and Control (CDC)}.\hskip 1em plus 0.5em minus 0.4em\relax IEEE, 2015, pp. 7034--7041. [Online]. Available: \url{https://doi.org/10.1109/CDC.2015.7403328}
\BIBentrySTDinterwordspacing

\bibitem{Budivsic2012}
\BIBentryALTinterwordspacing
M.~Budi{\v{s}}i{\'c}, R.~Mohr, and I.~Mezi{\'c}, ``{Applied Koopmanism},'' \emph{Chaos: An Interdisciplinary Journal of Nonlinear Science}, vol.~22, no.~4, 2012. [Online]. Available: \url{https://doi.org/10.1063/1.4772195}
\BIBentrySTDinterwordspacing

\bibitem{han2020deep}
Y.~Han, W.~Hao, and U.~Vaidya, ``{Deep Learning of Koopman Representation for Control},'' in \emph{2020 59th IEEE Conference on Decision and Control (CDC)}.\hskip 1em plus 0.5em minus 0.4em\relax IEEE, 2020, pp. 1890--1895.

\bibitem{huang2022convex}
B.~Huang and U.~Vaidya, ``{A Convex Approach to Data-Driven Optimal Control via {P}erron-{F}robenius and {K}oopman Operators},'' \emph{IEEE Transactions on Automatic Control}, 2022.

\bibitem{vaidya2023data}
U.~Vaidya and D.~Tellez-Castro, ``{Data-Driven Stochastic Optimal Control with Safety Constraints using Linear Transfer Operators},'' \emph{IEEE Transactions on Automatic Control}, 2023.

\bibitem{vaidya2024}
U.~Vaidya, ``{When Koopman Meets Hamilton and Jacobi},'' 11 2024.

\bibitem{mohr2014construction}
R.~Mohr and I.~Mezić, ``{Construction of Eigenfunctions for Scalar-type Operators via Laplace Averages with Connections to the Koopman Operator},'' 2014.

\bibitem{schmid2010dynamic}
P.~J. Schmid, ``{Dynamic Mode Decomposition of Numerical and Experimental Data},'' \emph{Journal of Fluid Mechanics}, vol. 656, pp. 5--28, 2010.

\bibitem{EDMD2015}
\BIBentryALTinterwordspacing
M.~O. Williams, I.~G. Kevrekidis, and C.~W. Rowley, ``{A Data-Driven Approximation of the Koopman Operator: Extending Dynamic Mode Decomposition},'' \emph{Journal of Nonlinear Science}, vol.~25, pp. 1307--1346, 2015. [Online]. Available: \url{https://doi.org/10.1007/s00332-015-9258-5}
\BIBentrySTDinterwordspacing

\bibitem{AjinkyaMECC2023}
\BIBentryALTinterwordspacing
A.~Joglekar, C.~Samak, T.~Samak, K.~C. Kosaraju, J.~Smereka, M.~Brudnak, D.~Gorsich, V.~Krovi, and U.~Vaidya, ``{Analytical Construction of Koopman EDMD Candidate Functions for Optimal Control of Ackermann-Steered Vehicles},'' \emph{IFAC-PapersOnLine}, vol.~56, no.~3, pp. 619--624, 2023, 3rd Modeling, Estimation and Control Conference MECC 2023. [Online]. Available: \url{https://doi.org/10.1016/j.ifacol.2023.12.093}
\BIBentrySTDinterwordspacing

\bibitem{AjinkyaIROS2023}
\BIBentryALTinterwordspacing
A.~Joglekar, S.~Sutavani, C.~Samak, T.~Samak, K.~Kosaraju, J.~Smereka, D.~Gorsich, U.~Vaidya, and V.~Krovi, ``{Data-Driven Modeling and Experimental Validation of Autonomous Vehicles Using Koopman Operator},'' in \emph{2023 IEEE/RSJ International Conference on Intelligent Robots and Systems (IROS)}, 10 2023, pp. 9442--9447. [Online]. Available: \url{https://doi.org/10.1109/IROS55552.2023.10341797}
\BIBentrySTDinterwordspacing

\bibitem{MMPK2024}
\BIBentryALTinterwordspacing
A.~Joglekar, C.~Samak, T.~Samak, V.~Krovi, and U.~Vaidya, ``{Expanding Autonomous Ground Vehicle Navigation Capabilities through a Multi-Model Parameterized Koopman Framework},'' 2024. [Online]. Available: \url{https://www.researchgate.net/publication/380152547}
\BIBentrySTDinterwordspacing

\bibitem{Ajinkya_Adaptive_MMPK}
A.~Joglekar, U.~Vaidya, and V.~Krovi, ``{Modeling and Control of Off-road Autonomous Vehicles with Situationally Aware Data-Driven Framework},'' August 2024.

\bibitem{Dolgov2008}
D.~Dolgov, S.~Thrun, M.~Montemerlo, and J.~Diebel, ``{Practical Search Techniques in Path Planning for Autonomous Driving},'' \emph{AAAI Workshop - Technical Report}, 01 2008.

\bibitem{AutoDRIVE-Ecosystem-2023}
\BIBentryALTinterwordspacing
T.~Samak, C.~Samak, S.~Kandhasamy, V.~Krovi, and M.~Xie, ``{AutoDRIVE: A Comprehensive, Flexible and Integrated Digital Twin Ecosystem for Autonomous Driving Research \& Education},'' \emph{Robotics}, vol.~12, no.~3, 2023. [Online]. Available: \url{https://doi.org/10.3390/robotics12030077}
\BIBentrySTDinterwordspacing

\bibitem{AutoDRIVE-Autoware-2024}
T.~V. Samak, C.~V. Samak, and V.~N. Krovi, ``{Towards Validation of Autonomous Vehicles Across Scales using an Integrated Digital Twin Framework},'' in \emph{2024 IEEE International Conference on Advanced Intelligent Mechatronics (AIM)}, 2024, pp. 1068--1075.

\bibitem{AutoDRIVE-Simulator-2021}
\BIBentryALTinterwordspacing
T.~V. Samak, C.~V. Samak, and M.~Xie, ``{AutoDRIVE Simulator: A Simulator for Scaled Autonomous Vehicle Research and Education},'' in \emph{2021 2nd International Conference on Control, Robotics and Intelligent System}, ser. CCRIS'21.\hskip 1em plus 0.5em minus 0.4em\relax New York, NY, USA: Association for Computing Machinery, 2021, p. 1–5. [Online]. Available: \url{https://doi.org/10.1145/3483845.3483846}
\BIBentrySTDinterwordspacing

\bibitem{Kazhdan2006}
\BIBentryALTinterwordspacing
M.~Kazhdan, M.~Bolitho, and H.~Hoppe, ``{Poisson Surface Reconstruction},'' in \emph{Proceedings of the Fourth Eurographics Symposium on Geometry Processing}, ser. SGP '06.\hskip 1em plus 0.5em minus 0.4em\relax Goslar, DEU: Eurographics Association, 2006, p. 61–70. [Online]. Available: \url{https://dl.acm.org/doi/10.5555/1281957.1281965}
\BIBentrySTDinterwordspacing

\end{thebibliography}

\end{document}